\documentclass{article}

\usepackage{PRIMEarxiv}
\usepackage[utf8]{inputenc} 
\usepackage[T1]{fontenc}    
\usepackage{hyperref}       
\usepackage{pifont}
\usepackage{url}            
\usepackage{booktabs}       
\usepackage{amsfonts}       
\usepackage{nicefrac}       
\usepackage{microtype}      
\usepackage{lipsum}
\usepackage{fancyhdr}       
\usepackage{graphicx}       
\usepackage{amsmath}
\usepackage{times}
\usepackage{multirow} 
\graphicspath{{media/}}     

\title{A Cost-effective, Stand-alone, and Real-time TinyML-Based Gait Diagnosis Unit Aimed at Lower-limb Robotic Prostheses and Exoskeletons 
}

\author{
  Zarin Anjum Madhiha \\
  Dept. of Electrical and Electronic Engineering \\
  Brac University \\
  Dhaka 1212, Bangladesh\\
  \texttt{zarin.anjum.madhiha@g.bracu.ac.bd} \\
      \And
  Antar Mazumder \\
  Dept. of Mechatronics Engineering \\
  Rajshahi University of Engineering \& Technology (RUET) \\
  Rajshahi 6204, Bangladesh\\
  \texttt{antar.mte@ieee.org} \\
   \And
  Sohani Munteha Hiam \\
  Dept. of Electrical and Electronic Engineering \\
  Rajshahi University of Engineering \& Technology (RUET) \\
  Rajshahi 6204, Bangladesh\\
  \texttt{sohani@eee.ruet.ac.bd} \\
}

\begin{document}
\maketitle

\begin{abstract}

Robotic prostheses and exoskeletons can do wonders compared to their non-robotic counterpart. However, in a cost-soaring world where 1 in every 10 patients has access to normal medical prostheses, access to advanced ones is, unfortunately, extremely limited especially due to their high cost, a significant portion of which is contributed to by the diagnosis and controlling units.  However, affordability is often not a major concern for developing such devices as with cost reduction, performance is also found to be deducted due to the cost vs. performance trade-off.  Considering the gravity of such circumstances,  the goal of this research was to propose an affordable wearable real-time gait diagnosis unit (GDU) aimed at robotic prostheses and exoskeletons. As a proof of concept, it has also developed the GDU prototype which leveraged TinyML to run two parallel quantized int8 models into an ESP32 NodeMCU development board (7.30 USD) to effectively classify five gait scenarios (idle, walk, run, hopping, and skip) and generate an anomaly score based on acceleration data received from two attached IMUs. The developed wearable gait diagnosis stand-alone unit could be fitted to any prosthesis or exoskeleton and could effectively classify the gait scenarios with an overall accuracy of 92\% and provide anomaly scores within 95-96 ms with only 3 seconds of gait data in real-time.

\end{abstract}

\keywords{Rehabilitation Robotics\and Robotic Prostheses\and Affordable Robotic Prostheses\and Robotic Exoskeletons\and Human Gait Classification\and TinyML\and Embedded AI}

\section{Introduction}
\textbf{\LARGE{F}}rom birth defects to unexpected accidents, since the dawn of the human race, people have been consistently exposed to the unforgiving consequences of losing their limbs leaving them incapable of standard human mobility. However, the untamable human nature to overcome imposed limitations has prompted them to fight back against the unfavourable consequences of losing limbs with various types of prostheses dating back to early Egyptian civilization in 950 B.C.~\cite{sitn}. Prostheses, as widely considered, are artificial limbs dedicated to mimicking or compensating the functionalities of the lost limbs to some partial extent. Robotic prostheses or powered prostheses are an upgrade to the traditional mechanical prostheses that harness the capabilities of modern electro-mechanical components and intelligent control systems to provide more accurate gaits~\cite{lenzi2014speed}. The concept of exoskeletons, on the other hand, originated with the goal of augmenting the pre-existing mobility state, either unimpaired or impaired. These devices are primarily dedicated to rehabilitation purposes such as allowing patients to walk after losing locomotion due to spinal cord injuries, stroke, or other causes, which do not necessarily require the absence of limbs, unlike prostheses. Modern robotic exoskeletons utilize mechanical or electro-mechanical components such as motors, hydraulic actuators, levers, and so on~\cite{gorgey2018robotic}. For a long time, the developments for these devices were only limited to mechanical advancements such as material improvements, joint improvements,  and weight-minimization which, albeit made them comfortable and easier to use, could not drastically increase the fidelity of these devices compared to the natural locomotion of the lost limb or human-body in general. The incorporation of modern electro-mechanical components transformed the landscape of medical research, ushering in a new era of robotic rehabilitation with improved prostheses and exoskeletons. The advent of robotic prostheses and exoskeletons in the domain of medical science has led to a much-needed paradigm shift in terms of the capability of these devices. However, despite the promises and the huge potential robotic prostheses and exoskeletons offer compared to their traditional counterpart, some hindrances such as the cost and complexity are rather amplified, making them less accessible~\cite{gorgey2018robotic}.


\begin{figure}
    \centering
    \includegraphics[width=1\linewidth]{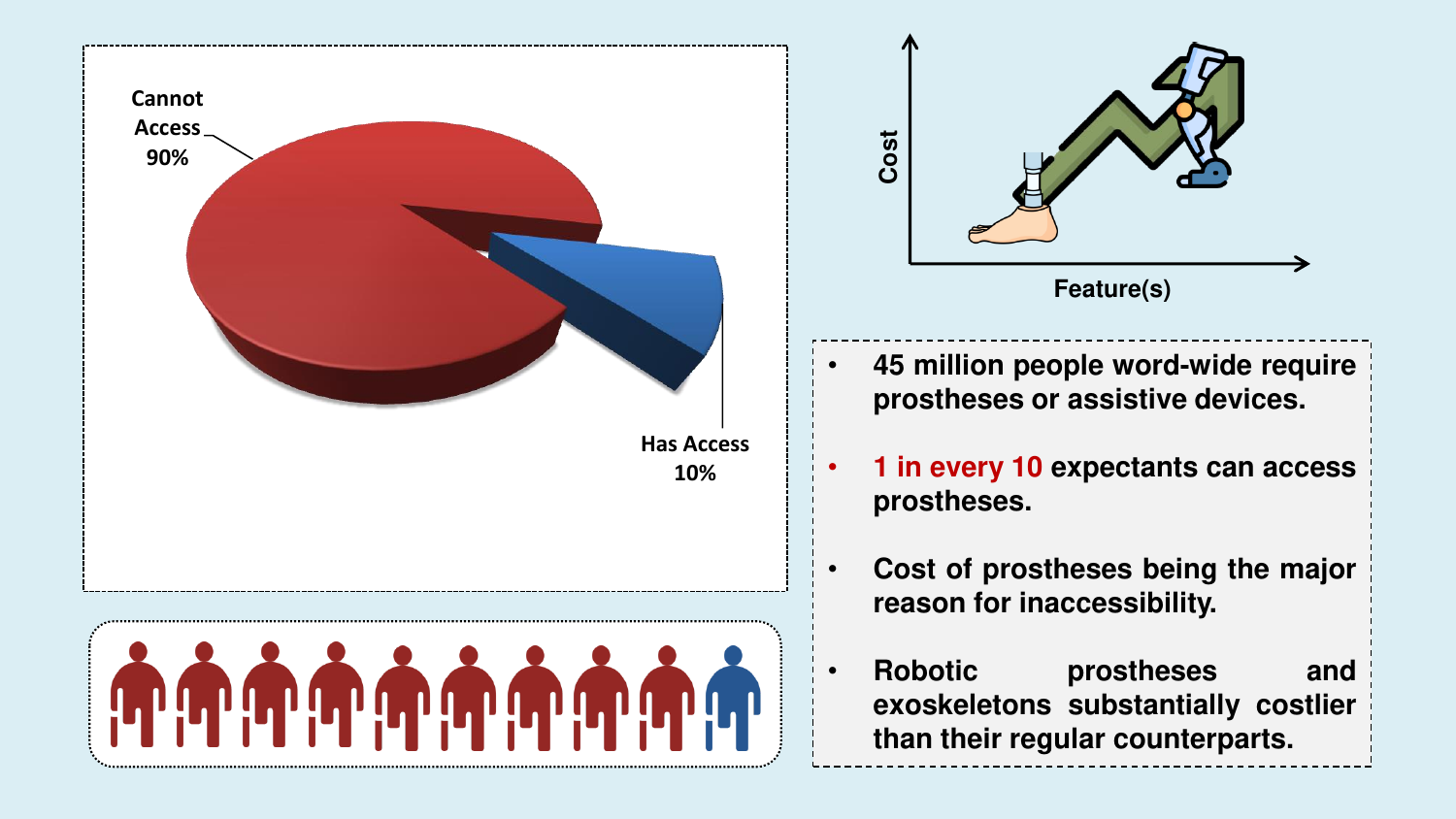}
    \caption{Disparity in accessibility to prostheses and assistive devices: Demand vs. Access and the major catalyst.}
    \label{disparity}
\end{figure}

\begin{figure}
    \centering
    \includegraphics[width=1\linewidth]{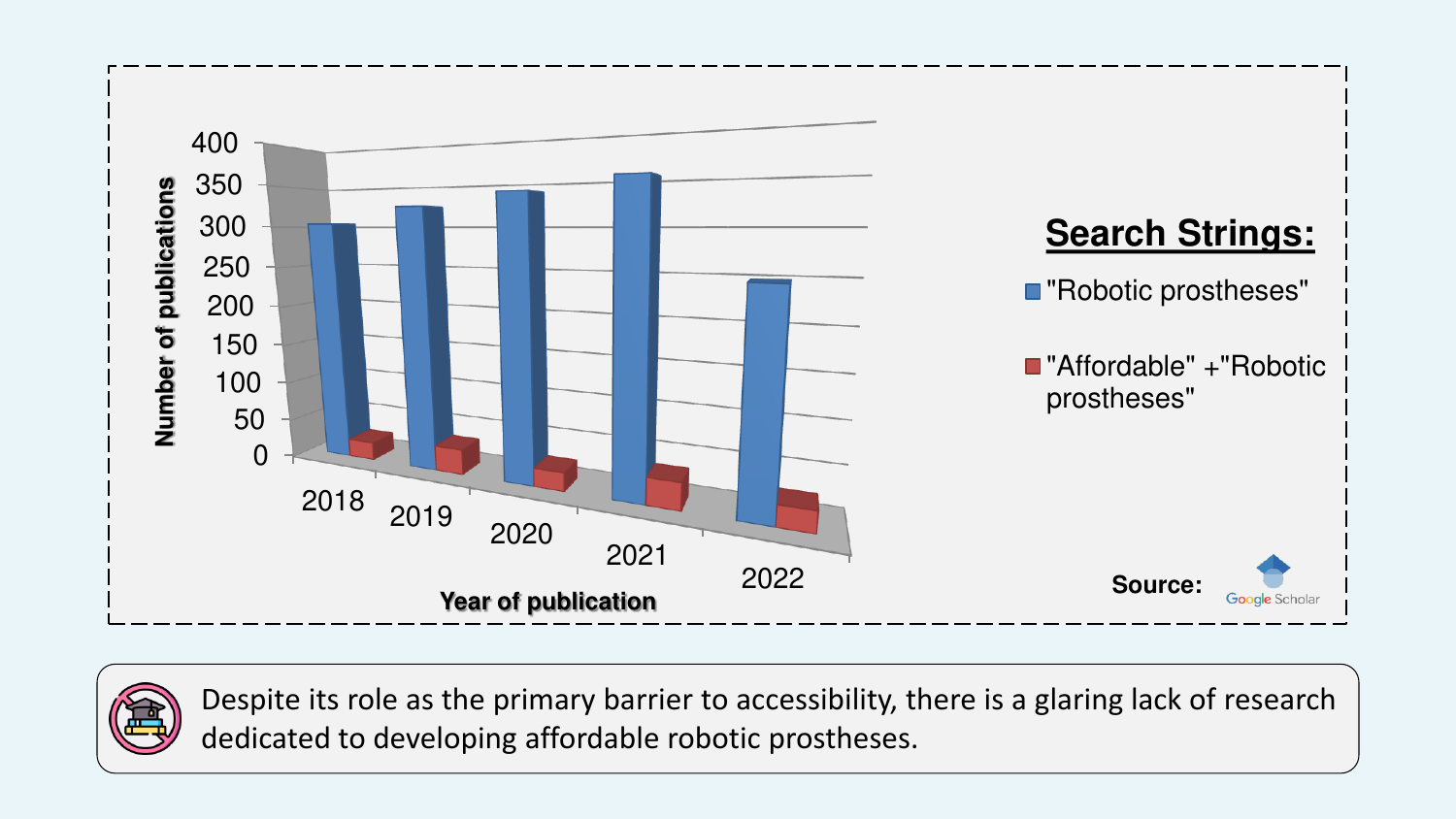}
    \caption{Disparity in research focus: Robotic Prostheses vs. Affordable Robotic Prostheses based on the yearly number of publications.}
    \label{lack}
\end{figure}

Summarized in figure~\ref{disparity}, a 2017 global estimate by the World Health Organization (WHO) stated that up to 40 million people were reported to have been in need of medical prostheses or assistive devices~\cite{world2017standards, chadwell2020technology} whereas, only 1 in every 10 people could access them due to their high cost and other impediments~\cite{world2017standards}. Although there is no strong numerical data dedicated to robotic prostheses and exoskeletons expressing such disparity, it is, for certain, much more severe as they are significantly costlier compared to their non-robotic counterpart. A major reason behind such high cost is the cost of the processing units leveraged for diagnosing the user gaits for any anomalies, and also for controlling the device. However, as evident in figure~\ref{lack}, although the cost is the primary impediment to making robotic prostheses accessible, there is an obvious lack of research focus on developing affordable robotic prostheses compared to the whole research domain dedicated to robotic prostheses hinted by the number of yearly publication indexed in the esteemed Google Scholar database. A common reason behind such a tendency is the trade-off between cost and performance as to alleviate the cost, less sophisticated equipment is to be used which is not adequate in a sensitive application such as medical prostheses where precision and performance are primary priorities.   Considering the gravity of such issues in hand, this paper intended to delve deep into a comprehensive literature review exploring the recent advancements in the field of wearable gait diagnosis systems which could be potentially applied to robotic prostheses and exoskeletons. It was found that the gait diagnosis units (GDUs) leveraged were not only expensive but also lacked stand-alone capabilities in general. To address the research gap, this study proposed a wearable cost-effective (7.30 USD) stand-alone GDU based on TinyML. The quantized int8 Artificial Neural Network (ANN) model could successfully classify five types of gait scenarios -\textit{Idle, Walk, Run, Hopping,} and \textit{Skip} with an overall classification accuracy of 92\%. Additionally, the device ran a K-means clustering model in parallel that generated an anomaly score for any unexpected movement scenarios. As a proof of concept, the wearable GDU was integrated with a lower-limb exoskeleton prototype where it provided inference within 95-96 ms after taking just 3 seconds of gait data.

Next, Section 2 goes through a literature review of related works providing a glimpse of the state-of-the-art and the research gaps. The following section describes the methodology including the design of the GDU, data collection, processing, the parallel model architecture, and model deployment. Section 4 delineates the results in terms of model performance, average anomaly score for various age groups, and latency calculation followed by a discussion highlighting the strengths and limitations of this research. Finally, the study is concluded in Section 6 and future plans are briefed.

\section{Related Works}

Gait classification has been mostly carried out by swing phase analysis~\cite{trivedi2024advances}. As the foot acceleration varies during these phases, accelerometer data has been repeatedly used for affordable yet effective gait detection. As acceleration data collected from IMUs are mostly time series data, methods like Long Short-Term Memory (LSTM) were frequently employed. For example, Tan \textit{et al.} leveraged accelerometer-based data collected from inexpensive wearable devices to train a modified LSTM model in~\cite{tan2019time} that could classify two gait events, namely Heel-strikes and Toe-offs for different types of gait activities. These activities were running and walking in both indoor and outdoor environments. The developed model was compared with the original LSTM model and six benchmark models for Gait Event Detection (GED) where the modified LSTM network outperformed them all exhibiting a median F1-score up to 0.98. However, despite an approach towards affordable GED, there was no hardware or physical implementation of this study. The same year, in \cite{zhen2019walking}, another GED method leveraged LSTM-DNN (Long Short-Term Memory - Deep Neural Network) algorithm to distinguish patterns in stance and swing phases. LSTM, K-Nearest Neighbour (KNN), and Support Vector Machine (SVM)  algorithms were compared with the LSTM-DNN where the latter showed the highest accuracy for gait phase detection, which was 91.8\%. However, the approach utilised a one-way Bluetooth connection to send the gait acceleration data over to a working computer where the gaits were analysed lacking a stand-alone feature. A recent approach in \cite{lee2021continuous}, demonstrated LSTM network gaining an impressive mean absolute error below 0.01 using gait data from two 9-axis Inertial Measurement Units (IMUs) across various walking speeds. However, there was no on-board real-time application as it had a unidirectional data flow where the data collected were analysed in an expensive processing set-up comprising an I7-9700 CPU and a NVIDIA RTX 2060m GPU with 6 GB VRAM. A novel labelling procedure using LSTM was propounded in \cite{hong2022piecewise} where the authors used a piecewise linear label for human gait phase estimation in three-speed conditions- slow, general and normal-fast. It exhibited notable improvement in estimation accuracy and speed adaptability, particularly during the mid-stance phase. Hong \textit{et al}. explored the viability of torso kinematics on gait phase estimation using a customised Artificial Neural Network (ANN) architecture with LSTM and Bi-LSTM layers. The approach achieved an MSE of 5.34E-04 ± 7.11E-04 in the mid-stance gait phase \cite{hong2022effect}.

Apart from LSTMs, other methods, such as Deep Convolutional Neural Network (DCNN), Backpropagation Neural Network (BPNN), and Graph Convolution Network (GCN) were also implemented. DCNN was used in \cite{su2020gait} leveraging data from 7 wearable IMUs to classify five different gait phases with around 97\% accuracy. The study constructed "spatial patterns" by arranging IMU inputs in particular sequences which could be then considered as images to be fed into the DCNN as image input tensors. However, there was no real-time implementation. In~\cite{gong2020bpnn}, Gong \textit{et al.} developed a real-time GED system with a high classification accuracy based on BPNN. The method achieved a high real-time recognition accuracy of 98.03\% with an impressive inference latency of 0.9 ms in detecting six locomotion modes. Despite having excellent real-time performance metrics, the approach utilised an expensive cRIO9082-based controller attached to an active pelvis orthosis (APO)  harness. Figueiredo \textit{et al.} proposed an inertial sensor system in~\cite{figueiredo2020wearable} utilising 7 IMUs(MPU6050) for real-time kinematic gait data analysis. Its performance was compared with MVN BIOMECH where it exhibited better performance metrics in stairs compared to ramps and flat terrains. It employed various ML and DL models for different joints such as hip, knee, and ankle joints where more robust results were found in the case of the knee joint with an \(R^{2}\) score of over 0.90. However, with seven IMUs, an ARM-based STM32F407VGT microcontroller along other hardware, it had room for improvements in terms of compactness and further cost reduction. In \cite{li2021real}, the GED was performed by attaching 3 wearable infrared (IR) distance sensors exhibiting a detection rate of 99.62\%. However, the average error rate was quite high reaching 34ms. Also, the study did not address how it tackled detection errors due to reflective surfaces or other IR sources which could corrupt the input. A GCN model for gait phase classification with an accuracy of 97.43\% was demonstrated in \cite{wu2021gait}, which was claimed to be higher than the state-of-the-art models from the Euclidean domain such as LSTM and DCNN, which had an average of 70\%-80\% accuracy. Nevertheless, it was quite an expensive approach regarding the hardware and processing units involved. In a recent academic study \cite{choi2023walking}, the authors introduced a Gait Phase Estimation Module (GPEM). The study demonstrated that the combined approach of the GPEM and a real-time algorithm resulted in a significant improvement in gait phase estimation. Specifically, deviations compared to time-based estimation and phase portrait process were reduced by 48\% and 48.29\% respectively. However, the online algorithm was not fit for offline tasks; thus, lacking stand-alone capabilities. A CNN-LSTM-CBAM approach was proposed in \cite{yang2023real} which achieved an evaluation index \(R^{2}\) of 0.9221 for enhancing GED and the control strategy of the wearable soft exosuits that can be reconstructed to complex and different environments, but the method came with significant development cost. In \cite{zhang2024actuator}, Zhang \textit{et al.} adopted a deep learning approach using LSTM for the gait phase estimation. This method detected with 94.60\% accuracy but with an expensive Teensy 4.1-based microcontroller.

\begin{figure}
    \centering
    \includegraphics[width=1\linewidth]{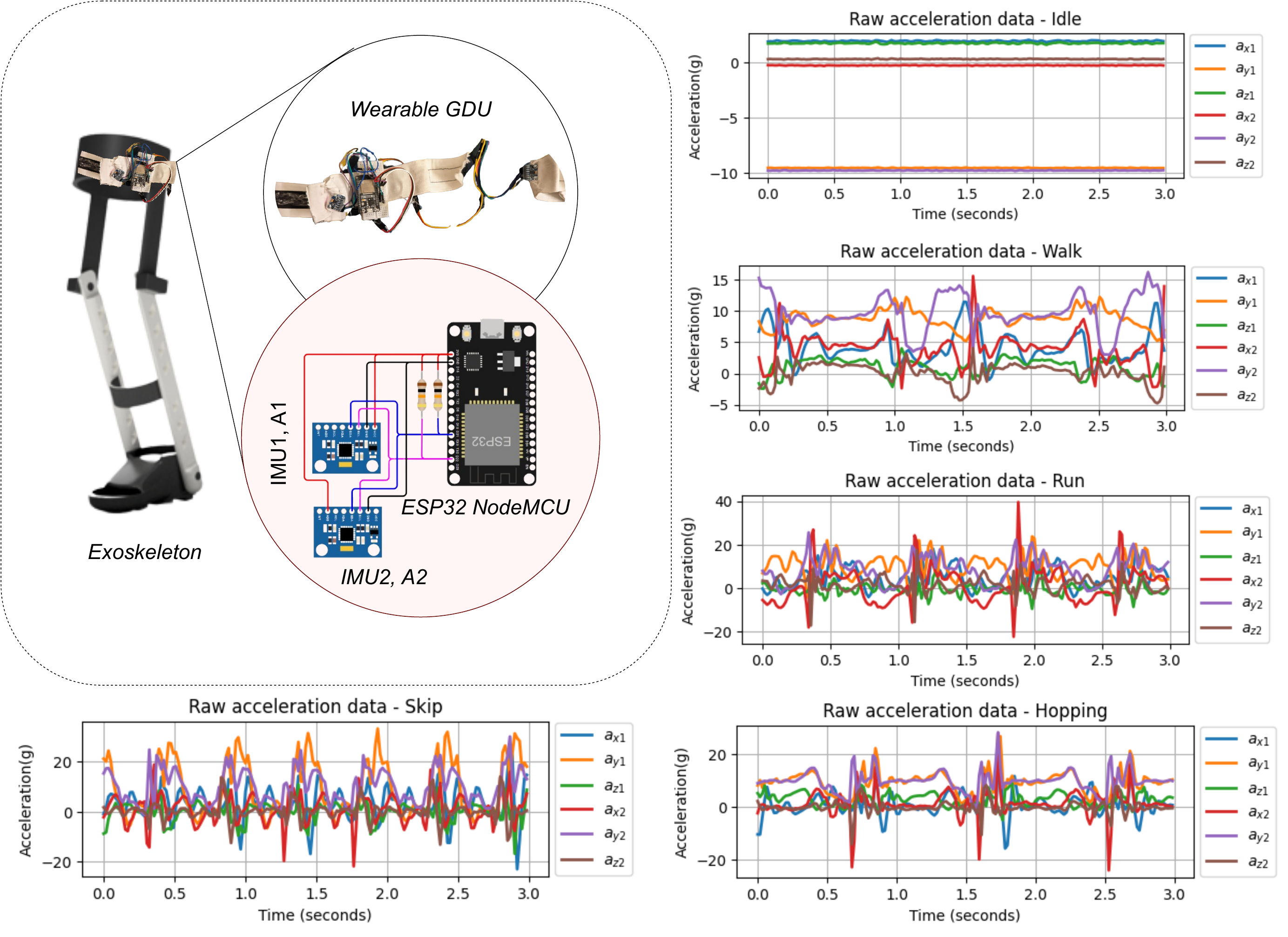}
    \caption{The GDU physical form and schematics and raw acceleration data from A1 and A2 for the five gait scenarios considered.}
    \label{fig:raw_data}
\end{figure} 


\begin{figure}
    \centering
    \includegraphics[width=1\linewidth]{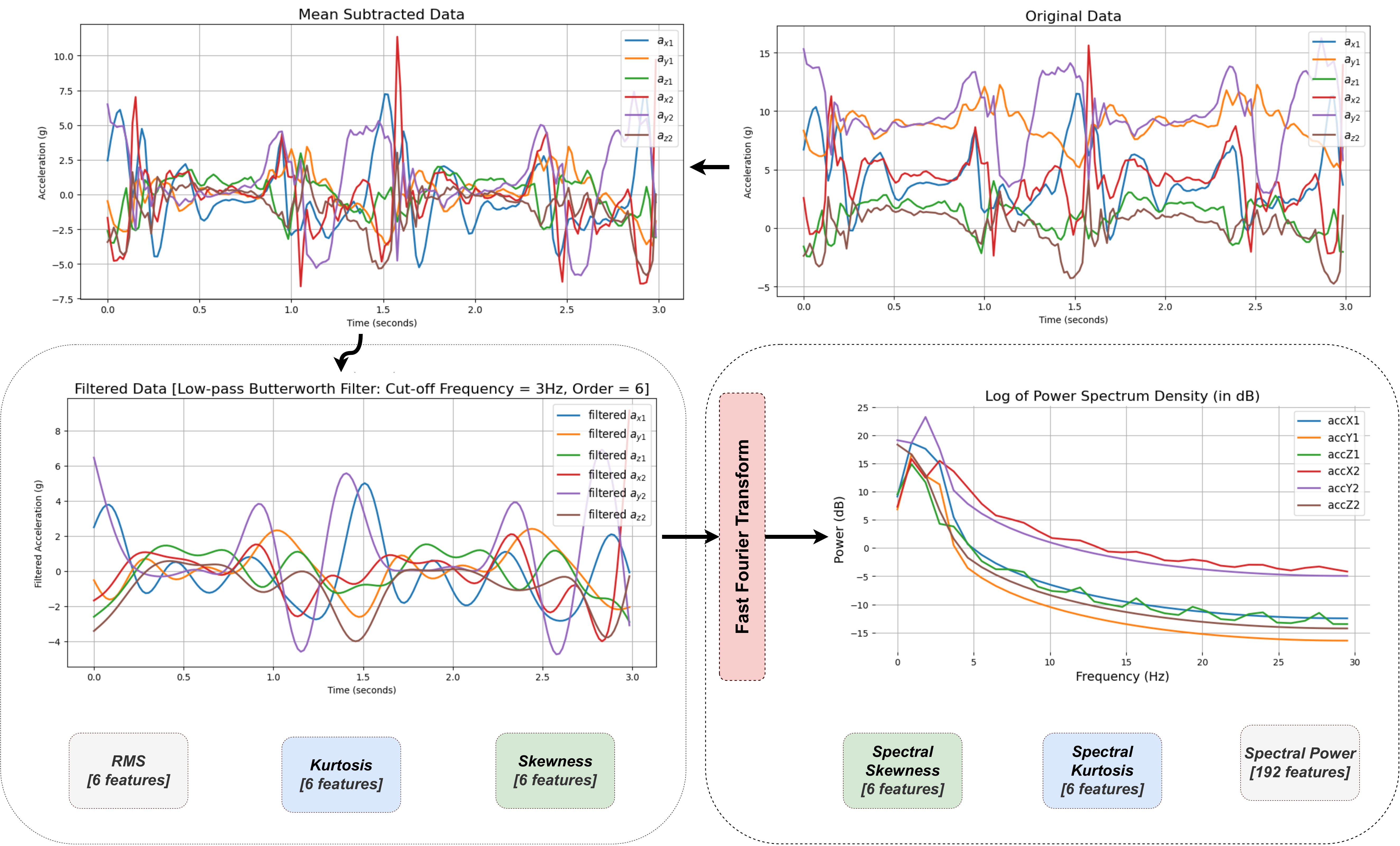}
    \caption{Gait data preprocessing and feature generation.}
    \label{fig:process}
\end{figure}

\section{Methodology}

There are two outputs to our approach- classified gait scenario and anomaly score. Gait scenarios comprise 5 different classes - walk, skip, run, idle, and hopping. Typically, most approaches recognise gait phases to detect gait anomalies. This study, on the other hand, recognises normal gait scenarios and then assigns an anomaly score to scenarios that deviate from the normal. The methodology adopted in this research can be divided into four phases - \textit{1) development of the gait diagnosis unit(GDU), 2) data collection and processing, 3) training of a modified CNN+K-means model,} and \textit{4) deployment in real-time gait diagnosis}.  

\subsection{Development of GDU}
The GDU, as demonstrated in figure~\ref{fig:raw_data}, is a wearable unit with two inexpensive six-axis MPU6050 accelerometers attached to an ESP32S 30P NodeMCU development board via two 1k Ohm pull-up resistors. The two accelerometers are used to send acceleration data to the MCU over I2C connections with addresses 0x68 and 0x69 respectively. One of the accelerometers, A1 is strapped to the mid-thigh segment of the user while the other one, A2 is attached to the ankle near the shank. These two have been found among the most optimal positions for gait classification~\cite{manna2023optimal}. The wearable unit can be placed on any exoskeleton or prosthesis due to its modular design. In this study, it was strapped to an aluminium-alloy exoskeleton prototype where a servo motor was utilised and controlled by the same MCU using PWM signals. The primary emphasis of this study was the development of an affordable GDU to minimize processing unit costs. Thus, \textit{it is imperative to acknowledge that the exoskeleton used in this study was an arbitrary choice}.

\subsection{Data Collection and Preprocessing}
Each of the accelerometers provided three acceleration values along the axes X, Y, and Z in 59 Hz. This resulted in the six primary inputs, which were labelled as \textbf{\(a_{x1}, a_{y1}, a_{z1}, a_{x2}, a_{y2},\)} and \(a_{z2}\) in units of g, where 1 g = 9.8 \(ms^{-2}\). These six parameters were recorded for all five gait scenarios among five volunteers between the ages of 21 to 55, two of whom were above the age of 30. Each candidate was instructed to perform the five gait scenarios with three minutes allotted to each gait scenario with varying speeds, resulting in 15 minutes of gait data for a single user for five gait scenarios. For example, figure~\ref{fig:raw_data} shows the raw acceleration pattern differences among different gait scenarios, where it is evident that some had more spikes, especially along the Z axis between the same time-frame. The gait data was fed into the edge impulse CLI framework via the data forwarder~\cite{edge_impulse_cli_2024}. For each axis for both A1 and A2, a 3-second window was considered to extract acceleration values along the time axis. The total number of data instances or points for each sample can be expressed as in equation~\ref{eq:sample_size}, where \(N_{T}\) is the total number of data points for each sample, \(w\) is the size of the window, \(c\) is the number of acceleration axes, and \(f\) is the frequency in Hz. Thus, for the 6 axes with a frequency of 59 Hz, a total of 1062 data points were collected for each gait scenario sample or 177 data points for each axis.

\begin{equation}
N_{T} = w \times c \times f
\label{eq:sample_size}
\end{equation}

\begin{equation}
\bar{a}_{ki} = \frac{1}{N}\sum_{n=1}^{N} a_{ki}(n)
\label{eq:mean_}
\end{equation}

\begin{equation}
a_{ki}(n) = a_{ki}(n)- \bar{a}_{ki}
\label{eq:mean_2}
\end{equation}

However, the raw data spikes as exhibited in figure \ref{fig:process} were inconsistent and also the influence of gravity had to be neutralised. So first, the mean value for each axis was subtracted, as expressed in equation~\ref{eq:mean_} and equation~\ref{eq:mean_2},  from individual data points to center the charts around zero-g. Here $a_{ki}(n)$ is the acceleration in g at the data point, n with k as the axis (x,y,z) and i as the IMU number(1,2) for each sample.  N is the total number of data points for each axis i.e., \(N = N_{T}/6\). ${\bar{a}_{ki}}$ is the mean acceleration for the gait scenario sample for all its data points in the 3-second window (i.e., N=177). Then a 6-order low-pass Butterworth filter with a cut-off frequency of 3 Hz was applied to get rid of the unreliable spikes in the higher frequency regions as demonstrated in the bottom rows of figure \ref{fig:process}. The filtered signal determined 3 types  of features \textit{for each sample} which were RMS (equation~\ref{eq: RMS}), Kurtosis (equation~\ref{eq: Kurtosis}), and Skewness \cite{schulte2022waveform} (equation~\ref{eq: skewness}). The kurtosis value used was Kurtosis excess~\cite{van2020empirical} with 3 subtracted so that for normal distribution the Kurtosis value would be zero. Since these terms are used interchangeably in practice, this study will refer to the feature as Kurtosis for ease.

\begin{equation}
 \text{\(a_{ki (RMS)}\)} = \sqrt{\frac{1}{N}\sum_{n=1}^{N} a_{ki}(n)^2}
\label{eq: RMS}
\end{equation}

\begin{equation}
a_{ki (Kurtosis)} = \frac{\frac{1}{N} \sum_{n=1}^{N} (a_{ki}(n) - \bar{a}_{ki})^4}{\left(\frac{1}{N} \sum_{n=1}^{N} (a_{ki}(n) - \bar{a}_{ki})^2\right)^2}- 3
\label{eq: Kurtosis}
\end{equation}

\begin{equation}
a_{ki(Skewness)} = \frac{\frac{1}{N} \sum_{n=1}^{N} (a_{ki}(n) - \bar{a}_{ki})^3}{\left(\frac{1}{N} \sum_{n=1}^{N} (a_{ki}(n) - \bar{a}_{ki})^2\right)^{3/2}}
\label{eq: skewness}
\end{equation}

3 types of spectral features were obtained \textit{for each sample} Spectral Skewness (equation~\ref{eq: Spectral skewness}), Spectral Kurtosis \cite{taylor2019spectral} (equation~\ref{eq: Spectral Kurtosis}), and Spectral Power \cite{vaseghi2000power} (equation~\ref{eq: Spectral Power}). Here \(K\) represents the frequency of the selected bin. Fast Fourier Transform (FFT) \cite{fu2020actuator} was performed (as in equation~ \ref{eq: FFT}) in the signal $a_{ki}(n)$ of 177 samples. The length of FFT, \(N_{L}\) was 64, which implies within the frequency range of 0-59 Hz, 64 bins had been considered resulting in \(64 \times 3\)   or 192 bins for the 3-second window which then provided 192 spectral power \cite{vaseghi2000power}(as in equation~\ref{eq: Spectral Power}) values used as features.

\begin{equation}
    A_{ki}(K) = \sum_{n=0}^{N-1} a_{ki}(n) 
    \cdot e^{-j\frac{2\pi Kn}{N}}
    \label{eq: FFT}
\end{equation}


\begin{equation}
A_{ki (Spectral~Skewness)} = \frac{\frac{1}{N} \sum_{k=1, i=1}^{N} |A_{ki}(K)|^3}{\left(\frac{1}{N} \sum_{k=1, i=1}^{N} |A_{ki}(K)|^2\right)^{3/2}}
\label{eq: Spectral skewness}
\end{equation}

\begin{equation}
A_{ki (Spectral~Kurtosis)} = \frac{\frac{1}{N} \sum_{k=1, i=1}^{N} |A_{ki}(K)|^4}{\left(\frac{1}{N} \sum_{k=1, i=1}^{N} |A_{ki}(K)|^2\right)^2} - 3
\label{eq: Spectral Kurtosis}
\end{equation}

\begin{equation}
    P_{ki (Spectral~ Power)} = \frac{1}{N_{L} \times f} \left| A_{ki}(K)\right|^2
    \label{eq: Spectral Power}
\end{equation}

\subsection{Model Architecture}
\begin{figure}
    \centering
    \includegraphics[width=1\linewidth]{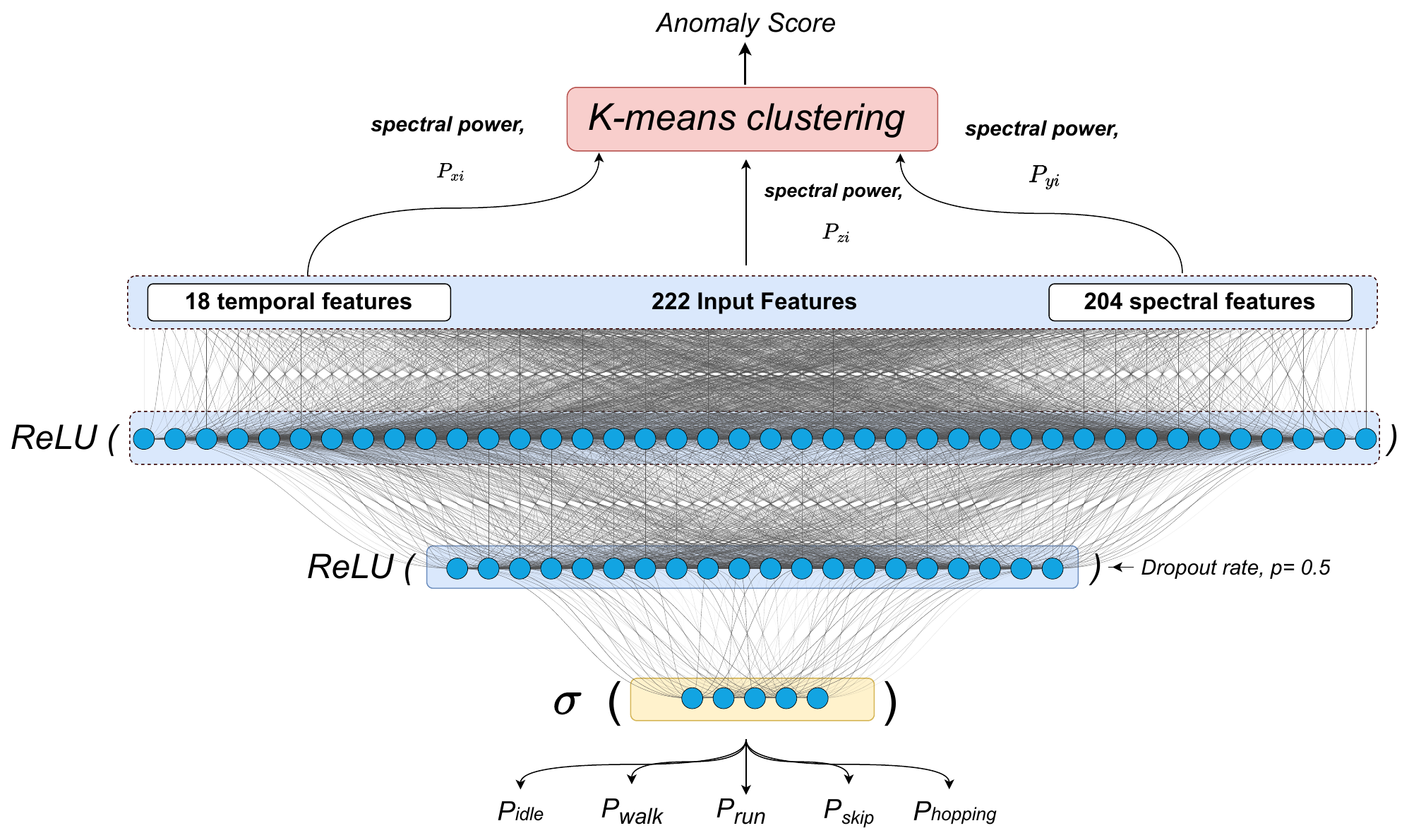}
    \caption{The parallel models leverage ANN to classify gait scenarios and K-means clustering to throw an anomaly score. }
    \label{fig:model_architecture}
\end{figure}

As evident in figure~\ref{fig:model_architecture}, this approach leveraged two models in parallel - an ANN to classify gait scenarios and a K-means clustering model for generating an anomaly score. 
First, the ANN model comprises the input layer with 222 nodes representing the 222 inputs. It is a simple structure with two fully connected hidden layers with \textit{ReLU} activation~\cite{agarap2018deep} and a dropout rate, \textit{p= 0.5} applied to the second hidden layer.  A Softmax activation~\cite{sharma2017activation} was applied to the output layer, which then provided the probability of the gait scenarios within a range (0,1). The shallow network was intended to reduce the computation load on the MCU adopted as it had limited resources. The training loop was executed for 500 epochs. After each forward pass, Categorical Cross Entropy (CCE) loss was calculated as equation ~\ref{eq:loss} where \(y_i\) is the true label probability and \(\hat{y}_i\) is the predicted label probability~\cite{martinez2018taming}. Adam optimizer\cite{kingma2014adam} was leveraged to adjust the weights with a learning rate of 0.0005 for training the network.

\begin{equation}
\text{Categorical Cross Entropy} = - \sum_{i=1}^{} y_i \cdot \log(\hat{y}_i)
\label{eq:loss}
\end{equation}


Simultaneously, four of the inputs (in this case, spectral powers: \(P_{x1}, P_{y1}, P_{z1}, P_{z2}\)) were taken and paired against each other along the X and Y axis denoted by the point (\(x_{test}\),\(y_{test}\)) on the XY plane. The selection of these features was performed inherently by the Ege Impulse studio based on its feature importance analysis~\cite{edgeimpulse-anomaly-detection}.  Now, based on which of the parameters were chosen, similar feature points with cluster size,\(N_{c}\) of 32 were considered in the XY plane. For example, if \(P_{x1}\) was taken along the X-axis and \(P_{z1}\) along the Y-axis, then 32 points, where the X-coordinates of the same belonged to \(P_{x1}\) and the Y-coordinates belonged to \(P_{z1}\) were considered. $A_{x}$ and $A_{y}$ represent these points of the cluster along the X and Y axes respectively. $C_{x}$ (as in equation~ \ref{eq:centroid A3}) and $C_{y}$ (as in equation~ \ref{eq:centroid A4}) are the X, Y coordinates of the centroid of the clustered points.

\begin{equation}
C_{x} = \frac{1}{N_{c}} \sum_{i=1}^{N_{c}} A_{xi}
\label{eq:centroid A3}
\end{equation}

\begin{equation}
C_{y} = \frac{1}{N_{c}} \sum_{i=1}^{N_{c}} A_{yi}
\label{eq:centroid A4}
\end{equation}

\begin{equation}
\text{Euclidean Distance~(d)} = \sqrt{{(C_{x} - x_{test})}^2 + {(C_{y} - y_{test})}^2}
\label{eq:Euclidean Distance}
\end{equation}

Next, the Euclidean distance \cite{aldino2021implementation} (equation~\ref{eq:Euclidean Distance}) was calculated from the centroid (\(C_{x}, C_{y}\)) to the test point (\(x_{test}, y_{test}\)). based on the combination of the four features, there were a minimum distance and a maximum distance, the average of these distances represented the anomaly score. In this study,  a tolerance of 3 was initially considered as the threshold. Therefore, any gait sample producing a higher score was considered as an anomaly.

\subsection{Deployment in Real-time Gait Diagnosis}
The original model developed was a float32 model. However, such a model, although typically used in most applications, is not suitable to run in a resource-limited development board such as the ESP32 NodeMCU. Therefore, the original model was first downgraded or quantized to an int8 profile \cite{schizas2022tinyml} and then turned into an Arduino library based on c-array via the Edge Impulse studio framework. The quantization process reduced the number of bits of the weights and biases from 32-bit to 8-bit. Then using the library in the Arduino IDE environment, the model embedded with the necessary conditions was uploaded to the ESP32s NodeMCU. The GDU was attached to the designated areas of the lower limb on top of the exoskeleton prototype and the users were instructed to perform various gait scenarios. First, the output of the model as well as latency values were displayed using a serial monitor. Later the serial monitor was replaced with a 0.96" I2C OLED display. In the later cases, the latency information was excluded and only the gait classification information was displayed to utilise the small display size.

\section{Results}

\begin{figure}
    \centering
    \includegraphics[width=0.75\linewidth]{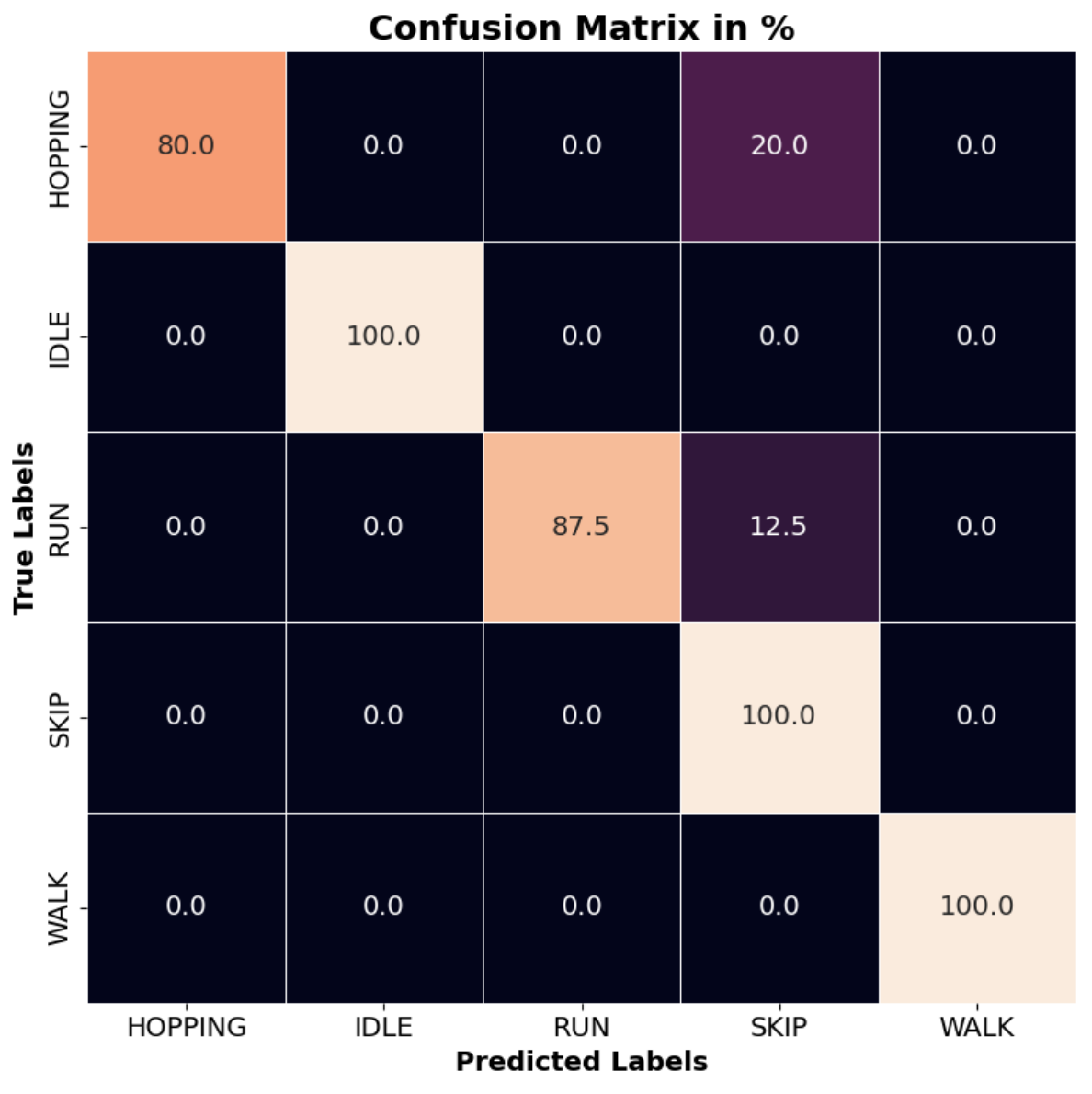}
    \caption{Confusion matrix for the Neural Network-based classifier.}
    \label{fig:cmat}
\end{figure}

There were three criteria for result analysis- 1) Performance of the model (precision, recall, accuracy, accuracy, f1-score, and average anomaly score), 2) Cost vs. performance trade-off, and 3) Data acquisition and inference latencies. There was no specific performance parameter for the k-means-based anomaly detection as the anomaly threshold was arbitrary and subject to a variable for different users. Nevertheless, the max value of this threshold was kept below 3.0.

\subsection{Performance Metrics}
In Figure \ref{fig:cmat}, the provided confusion matrix represents the performance of a classification model with five classes: Hopping, Idle, Run, Skip, and Walk. Each row represents the actual class, while each column represents the predicted class. The diagonal elements (from top-left to bottom-right) show that the majority of instances have been correctly classified, with a perfect 100\% accuracy for the Idle, Skip, and Walk classes. However, the classes, "Hopping" and "Run" have been predicted as "Skip" in around 20\% and 12.5\% of instances respectively, indicating some misclassification between these three classes. Overall, the model demonstrates near-perfect performance for most classes, resulting in an overall accuracy of 92\%. More details have been summed up in table~\ref{precision, recall, f1}, where it can be observed that all classes demonstrated high-performance metric values with a slight exception in the case of the "Hopping" and "Run" classes. 



\begin{table}
\caption{Performance Metrics for Precision, Recall, and F1 Score Across Different Classes.}
\label{precision, recall, f1}
\centering
\begin{tabular}{lcccccc}
\toprule
\textbf{Class} & \textbf{Precision} & \textbf{Recall} & \textbf{F1-Score} & \textbf{Support} \\
\midrule
HOPPING & 1.00 & 0.80 & 0.89 & 10 \\
IDLE & 1.00 & 1.00 & 1.00 & 7 \\
RUN & 1.00 & 0.88 & 0.93 & 8 \\
SKIP & 0.70 & 1.00 & 0.82 & 7 \\
WALK & 1.00 & 1.00 & 1.00 & 7 \\
\midrule
\textbf{Accuracy} &   &  & 0.92 & \\
\textbf{Macro avg} & 0.94 & 0.93 & 0.93 & 39 \\
\textbf{Weighted avg} & 0.95 & 0.92 & 0.93 & 39 \\
\bottomrule
\end{tabular}

\end{table}

\begin{figure}
    \centering
    \includegraphics[width=0.75\linewidth]{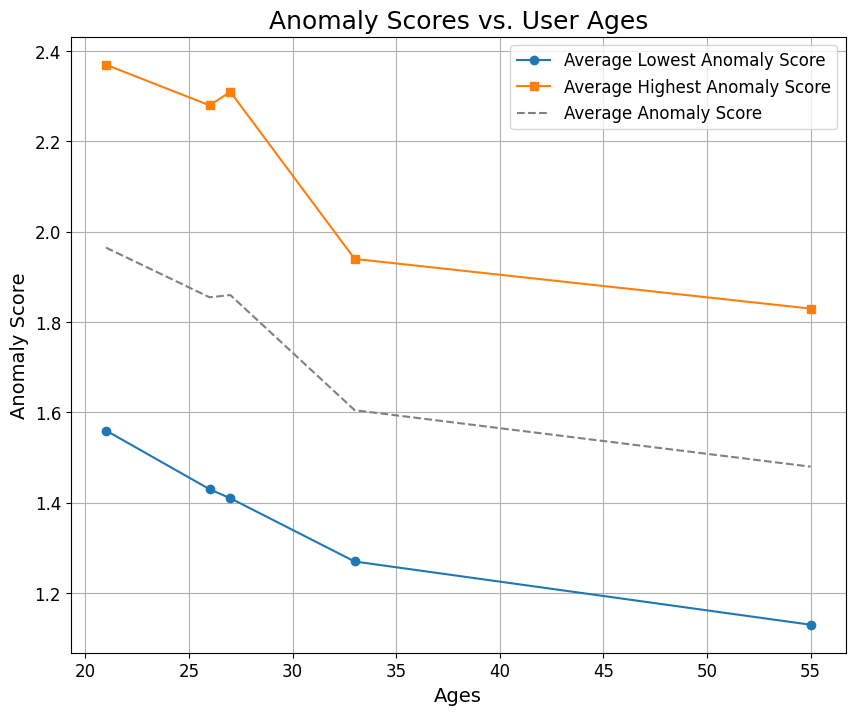}
    \caption{Average anomaly scores for users from different age groups: Higher margin visible among the youth.}
    \label{fig:anomaly score}
\end{figure}

As for the K-means-based anomaly score, it was found that the average anomaly scores, especially the average highest anomaly scores varied across different age groups. As demonstrated in figure~\ref{fig:anomaly score}, there was a clear decline in the average anomaly score among the older age groups compared to the youth. The highest average maximum anomaly value was found to be around 2.38, which was well below the set threshold of 3.

\subsection{Cost vs Performance Trade-off Comparison with Related Approaches}

As evident from the table~\ref{tab:gait_detection_studies_accuracy_cost}, most similar approaches have leveraged expensive hardware to process gait data. These hardware components were either full-fledged desktop or laptop computers~\cite{zhen2019walking,su2020gait,islam2022real} that received gait data via Bluetooth or other sensors in real time. Otherwise, controllers with high computation abilities ~\cite{gong2020bpnn} were utilized for providing wearable stand-alone features. Despite having very high classification accuracy, these hardware components even came at over 1,000 USD, which was quite expensive. The closest competitor to our approach was found in~\cite{negi2022standalone} with a  97.8\% classification accuracy, which leveraged an Arduino Nano33 BLE development board to run real-time inference. However, our approach did the same with only 5.8\% less accuracy but almost at one-sixth of the cost. Furthermore, the float32 version of the model attained a similarly high accuracy score as the top-performing ones.

\begin{table}
    \centering
    \caption{Comparison based on cost vs performance trade-off\textit{ [Note: The costs of the equipment used in developing GDUs were estimated based on the contemporary pricing of the year 2023. These prices are subject to change with time.]}.}
    \label{tab:gait_detection_studies_accuracy_cost}
    \begin{tabular}{cccccc}
        \toprule
        \textbf{Reference} & \textbf{Year} & \textbf{Gait Diagnosis Unit} & \textbf{Classification Accuracy (\%)} & \textbf{Cost (USD)} & \textbf{Stand Alone} \\
        \midrule

        Zhen \textit{et al.} \cite{zhen2019walking} & 2019 & PC/Work Station & 91.8 & 500+ & \ding{55} \\
        Gong \textit{et al.} \cite{gong2020bpnn} & 2020 & cRIO9082-based Controller & 98.03 & 1,000+ & \checkmark \\
       
        Su \textit{et al.} \cite{su2020gait} & 2020 & Lenovo Thinkpad T470p & 97 & 500+ & \ding{55} \\
        Negi \textit{et al.}\cite{negi2022standalone}& 2021& Arduino Nano 33 BLE & 97.8 & 40.50 & \checkmark \\
        Isam \textit{et al.}\cite{islam2022real}& 2022& PC/Work Station & 97.15 & 500+ & \ding{55} \\
         Zhang \textit{et al.}\cite{zhang2024actuator}& 2024& Teensy 4.1 & 94.60 & 94 & \ding{55} \\
        \textbf{This Study}& \textbf{---} & \textbf{ESP32s NodeMCU}  & \textbf{92 (int8), 97.14 (float32)} & 7.30 & \checkmark \\
        \bottomrule
    \end{tabular}
\end{table}

\subsection{Latency Analysis}

\begin{figure}
    \centering
    \includegraphics[width=1\linewidth]{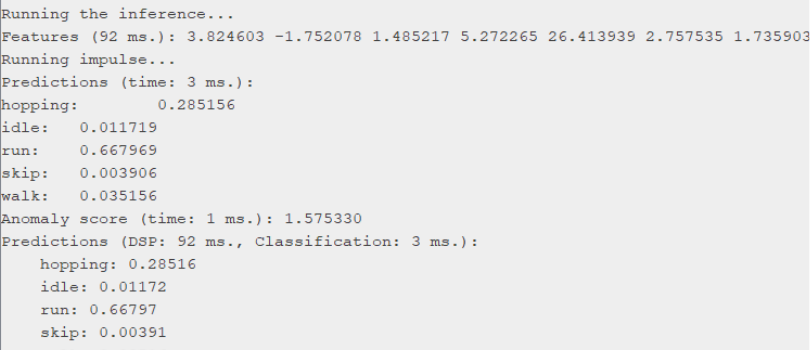}
    \caption{Classification results and corresponding latency breakdown after data collection derived from the Arduino IDE serial monitor.}
    \label{fig:latency}
\end{figure}

After the 3-second data collection window, latency has been distributed to three applications - feature generation, classification and anomaly score generation. 
From the serial monitor output taken from the Arduino IDE as illustrated in figure~\ref{fig:latency}, it can be observed that for our system took 92 ms for feature generation, 3 ms for gait scenario classification and 0-1 ms for generating the anomaly score. Thus, this approach could successfully detect gait scenarios along with gait anomalies within 95-96 ms after taking 3 sec of gait data.

\section{Discussion}

The proposed method took a different approach where it recognized the most common gait scenarios instead of gait events, unlike most contemporary studies. It provided a simple alternative to rigorous analysis based on gait events from which gait anomaly is typically derived. However, some limitations should be mentioned.
The first apparent downside of the model is its slightly lower classification accuracy compared to the hardware-intensive processes. Nevertheless, given that it was a quantized int8 model derived from a float32 variant to make it run in a resource-scare platform such as Esp32 NodeMCU with minimal latency, the accuracy reduction is understandable. Another concern can be considered regarding the absence of an algorithmic threshold determination for the K-means-based anomaly score. However, as found in the results, different people from different age groups would required different thresholds as the youth were able to perform more frequent and accelerated movements. However, considering the threshold was determined by trial and error, a more method-based process would be better.

Despite the shortcomings, given its stellar modularity, low training data requirement of only 15 minutes of gait data for each individual, and the feasibility to rapidly device custom GDUs for individual patients, the benefits far outweighed the limitations. Most importantly, with only a 4-6\% accuracy reduction compared to the state-of-the-art models, the GDU diagnoses user gait at a fraction of the cost, which makes it a viable step towards affordable robotic prostheses and exoskeleton development.

\section{Conclusion}
Affordability is a major concern for patients requiring advanced prostheses or exoskeletal systems. However, due to the cost vs. performance trade-off, it was observed to be a challenging task to develop high-accuracy gait diagnosis devices that are stand-alone, affordable and can perform in real-time. To address this issue, this research proposed and, as a proof of concept, developed a  TinyML-based wearable GDU that can perform gait scenario classification with an overall 92\% accuracy and generate a gait anomaly score based on only two IMUs and an inexpensive ESP32 NodeMCU development board which exhibited potential solution at a fraction of the state-of-the-art cost margins. The authors fervently believe that despite the minor limitations, this is the right step towards making robotic prostheses and exoskeletons more available to those in need. In the future, it is intended to increase the classification accuracy even higher and run inference on a smaller data collection window. It is also intended to use the IoT features of the NodeMCU to seamlessly integrate the exoskeleton model into a digital twin.

\section*{Ethics and Consent}
The five volunteers who participated in this experiment provided full verbal and written consent to contribute their gait data for the study. The research was approved by the Research \& Extension Committee of Rajshahi University of Engineering \& Technology, which was the ethical committee of the institute. No minor was part of this research.


\bibliographystyle{ieeetr}  
\bibliography{templateArxiv}

\end{document}